\pdfoutput=1

\documentclass[11pt]{article}

\usepackage[final]{acl}

\usepackage{times}
\usepackage{latexsym}
\usepackage{algorithm}
\usepackage{algorithmic}
\usepackage{amsmath}
\usepackage{booktabs}
\usepackage{tabularx}
\usepackage{multirow}
\usepackage{subfigure}
\usepackage{stfloats}

\usepackage[T1]{fontenc}

\usepackage[utf8]{inputenc}

\usepackage{microtype}

\usepackage{inconsolata}

\usepackage{graphicx}

\title{Filter-And-Refine: A MLLM Based Cascade System for Industrial-Scale Video Content Moderation}

\author{
    {\bf Zixuan Wang}\thanks{Equal contribution.}, {\bf Jinghao Shi}\footnotemark[1] \\
    {\bf Hanzhong Liang}, {\bf Xiang Shen}, {\bf Vera Wen}, {\bf Zhiqian Chen}, {\bf Yifan Wu} \\
    {\bf Zhixin Zhang}, {\bf Hongyu Xiong}\\
    TikTok \\
    \texttt{\{zixuan.wang1, jinghao.shi\}@tiktok.com}
}

\begin{document}
\maketitle
\begin{abstract}
Effective content moderation is essential for video platforms to safeguard user experience and uphold community standards. 
While traditional video classification models effectively handle well-defined moderation tasks, they struggle with complicated scenarios such as implicit harmful content and contextual ambiguity. Multimodal large language models (MLLMs) offer a promising solution to these limitations with their superior cross-modal reasoning and contextual understanding. However, two key challenges hinder their industrial adoption. First, the high computational cost of MLLMs makes full-scale deployment impractical. Second, adapting generative models for discriminative classification remains an open research problem. In this paper, we first introduce an efficient method to transform a generative MLLM into a multimodal classifier using minimal discriminative training data. To enable industry-scale deployment, we then propose a router-ranking cascade system that integrates MLLMs with a lightweight router model. 
Offline experiments demonstrate that our MLLM-based approach improves F1 score by 66.50\% over traditional classifiers while requiring only 2\% of the fine-tuning data. Online evaluations show that our system increases automatic content moderation volume by 41\%, while the cascading deployment reduces computational cost to only 1.5\% of direct full-scale deployment. 

\end{abstract}

\begin{figure*}[t]
\centering
  \includegraphics[width=\textwidth]{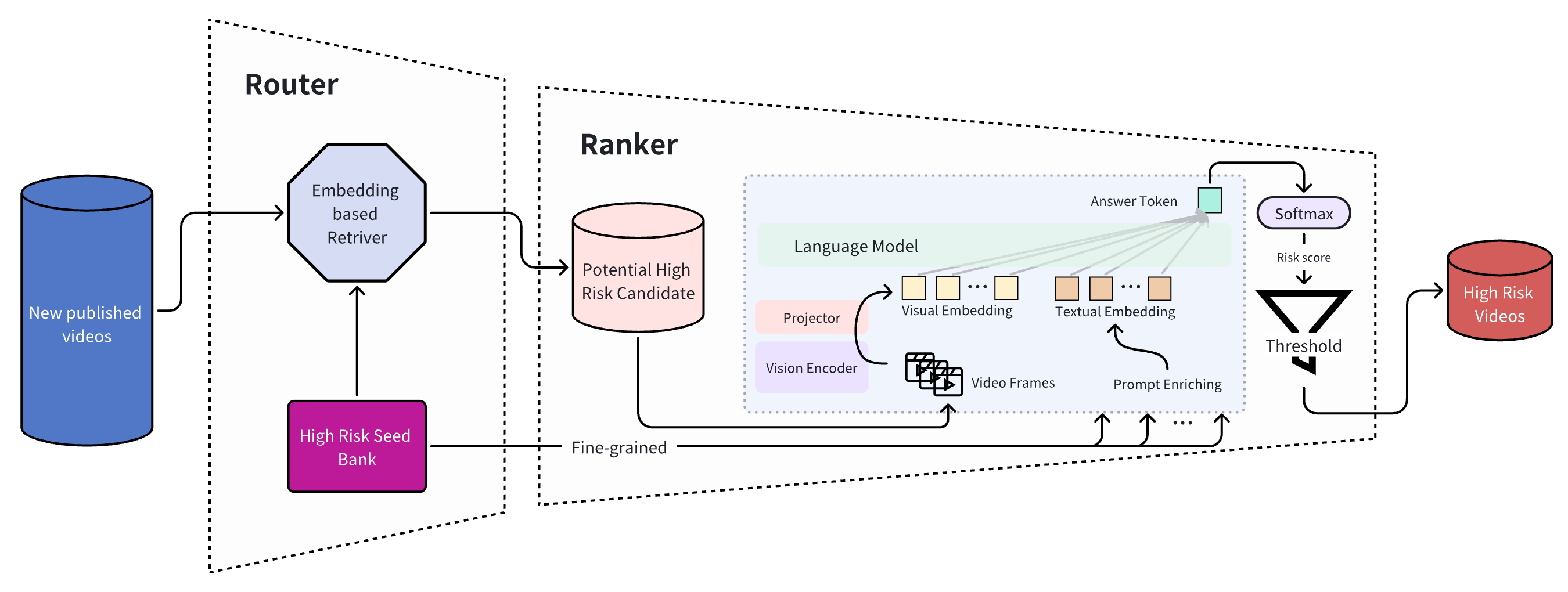}
  \caption{Overview of the cascade system design. The system consists of two stages: a Router and a Ranker. The Router filters and selects potentially high-risk content, while the Ranker performs fine-grained classification to refine the final decision.}
  \label{fig:system}
\end{figure*}

\section{Introduction}
The rapid expansion of short video platforms such as YouTube Shorts and Instagram Reels has transformed online content consumption. 
As user engagement and content volume continue to grow massively, effective content moderation has become more and more important.

Content moderation generally falls into two categories: human moderation and machine-driven auto-moderation. While human moderation provides good judgment, it is inherently slow, expensive, and difficult to scale.
As a result, machine learning (ML)-based auto-moderation has become crucial, offering scalable and efficient solutions for content moderation.

Currently, video content moderation is mostly handled by video classification models \cite{shi2024cpfd}, which process video inputs and tag videos based on a predefined taxonomy. While traditional video classification models effectively handle well-defined moderation tasks, they struggle with more complicated and context-dependent moderation challenges. For instance, they can reliably flag explicit harmful content but often fail to recognize implicit violations, such as subtle forms of misinformation or suggestive imagery. 
Multimodal Large Language Models (MLLMs) can be a promising alternative due to their superior reasoning and contextual understanding capabilities. 

Despite the potential of MLLMs in content moderation, two key challenges make their industrial deployment difficult. First, the high computational cost of large-scale MLLMs poses a big barrier for real industry deployment. 
To enable scalable deployment, we introduce a router-ranking cascade system. Inspired by recall-ranking architectures commonly used in recommendation systems, our approach employs a lightweight router as a first-stage filter. 
The router selectively passes only high-risk content, allowing the MLLM to focus on a small subset of potentially violating videos. 
The cascade design greatly reduces computational costs compared to direct full-scale deployment.

Second, as generative models, MLLMs are not inherently suited for discriminative classification tasks. Effectively converting a generative model into a classifier remains an open research problem. 
Some prior works \cite{chen2024mllm,mitra2025sparseattentionvectorsgenerative} have explored innovative approaches to this transformation, yet no existing study has specifically focused on adapting generative MLLMs for content moderation. In this paper, we address this gap with a straightforward yet effective transformation method that requires only minimal fine-tuning data while demonstrating strong performance in real-world content moderation applications.

We summarize our contribution as follows: 
\begin{itemize}
    \item 
    To the best of our knowledge, this work represents one of the first successful applications of MLLMs in a large-scale content moderation system.
    \item 
    We introduce a novel router-ranking cascade architecture that enables full-traffic deployment while significantly reducing computational costs.
    \item 
    We propose a straightforward yet effective method to adapt generative MLLMs for classification tasks, requiring only minimal fine-tuning data.
    \item 
    We validate the approach through comprehensive offline and online experiments on production data and enable deployment of the model in a real-world production environment. 
\end{itemize}

\section{Related Work}
\subsection{ML-based Content Moderation}
As social media platforms continue to expand, efficient content moderation becomes increasingly critical. Over the past few years, significant strides have been made in identifying harmful content such as hate speech \cite{das2023hatemm}, explicit material \cite{nudenet}, and toxic language \cite{yuan2024rethinking} across multiple modalities. Given that social media content naturally integrates video, images, and text, multimodal frameworks, for example \cite{yuan2024rethinking,binh2022samba}, have become a standard approach. Despite some relying on user feedback\cite{yu2025usmunbiasedsurveymodeling}, relatively few studies \cite{ye-etal-2023-multilingual,mullick2023content} focus solely on moderating images or text. With the rapid advancement of multimodal large language models (MLLMs), these techniques are increasingly being applied to content moderation, demonstrating strong performance \cite{ma2023adapting,wu2024icm}.

\subsection{MLLM and Supervised Fine-tuning}
Although Multimodal Large Language Models, such as LLaVA series \cite{NEURIPS2023_6dcf277e,Liu_2024_CVPR}, GPT-4 \cite{openai2024gpt4technicalreport} and DeepSeek series \cite{deepseekai2025deepseekv3technicalreport,deepseekai2025deepseekr1incentivizingreasoningcapability}, have shown versatility across diverse tasks, fine-tuning remains essential to achieve optimal performance for specific applications. InstructGPT \cite{NEURIPS2022_b1efde53} has demonstrated that with the help of human feedback, fine-tuning LLMs using reinforcement
learning from human feedback (\citet{learningtosummaize, deepreinforcement}) is able to outperform larger models. Furthermore, there are other parameter-efficient ways to leverage multimodal data, such as PEFT\cite{zhou-etal-2024-empirical} and FedMLLM\cite{xu2025fedmllmfederatedfinetuningmllm}.
The composition and quantity of data also significantly affect the capabilities of LLMs. \citet{dong-etal-2024-abilities, pareja2024unveilingsecretrecipeguide, pang-etal-2024-phased} highlight the need for strategic data selection and stages in the fine-tuning process to balance and optimize various model capabilities.
\\With the nature of generative models, MLLM does not demonstrate a strong capability in multimodal classification \cite{zhang2024visually}. \citet{chen2025can,liu2024agentps} explores the application in anomaly detection with different prompt formats.

\section{Cascade System Design}
Deploying Multimodal Large Language Models (MLLMs) at an industrial scale presents computational challenges, particularly for high-traffic platforms, where hundreds of millions of new videos are uploaded daily. Directly applying MLLMs to full traffic is prohibitively expensive and inefficient, which makes a scalable and resource-efficient moderation pipeline important.
Inspired by recall-ranking architectures in recommendation systems, we introduce a two-stage router-ranking cascade system in Figure~\ref{fig:system} to optimize moderation efficiency. This framework includes:

\texttt{Lightweight Router (Recall Stage).} A computationally efficient model acts as a first-stage filter, quickly identifying suspicious content while discarding low-risk videos.

\texttt{MLLM-Based Ranker (Ranking Stage).} The more powerful yet costly MLLM then analyzes only the high-risk subset, performing fine-grained reasoning to accurately detect harmful content.
\\This hierarchical filtering approach significantly reduces unnecessary MLLM processing, improving scalability while preserving high moderation accuracy on the real-time video platform.

\subsection{Router}
The router model serves as the first-stage filter in our cascade system\cite{liang2025embedding}. It can be implemented using any feasible architecture, such as classification models or embedding-based retrieval systems.
\\In our implementation, we leverage an embedding retrieval system as the router due to its effectiveness and efficiency. This system operates by maintaining a pre-selected bank of high-risk representative videos, called seed videos. The newly published videos are then filtered based on semantic similarity with the seed videos to pick high-risk candidates.
\\We designed several strategies to ensure high-quality seed selection, such as Centroid-Proximity Seed Selection, which uses clustering algorithms to identify good seeds, and Manual Seed Selection, which relies on annotators to identify "golden seeds". Our retrieval-based router offers several key advantages: Unlike classification models, our approach does not require labeled data and is trained in an unsupervised manner. The seed bank architecture offers the system rapid adaptation and great flexibility. By efficiently filtering content before MLLM processing, our router significantly reduces computational costs while maintaining high recall for potentially violating videos.

\subsection{Ranker}
The MLLM serves as the ranker, refining the Router’s output by predicting a more precise moderation decision. It takes both the extracted visual features from the video and a task-specific prompt corresponding to the target class. The model outputs a single token representing the predicted label and token probabilities as the confidence score. Unlike conventional classifiers with fixed output structures, MLLMs offer greater flexibility through prompt engineering, enabling adaptation to various moderation tasks without retraining. Their advanced reasoning and contextual understanding further enhance ranking performance, allowing the model to act as a strong refiner in the cascade system. Additionally, the extensive pretraining on open-domain knowledge provides a strong initialization for the ranking stage. For details on the MLLM-based ranker, refer to the next section.

\section{Finetune MLLM as Ranker}
In this section, we first introduce the multimodal large language model (MLLM) architecture. We then describe our continuous supervised fine-tuning process, covering the construction of the fine-tuning dataset and two fine-tuning strategies explored to optimize the model's performance. Next, we outline how the model's output is calibrated into probabilistic scores for online serving. Finally, we discuss further improvements such as prompt engineering and result ensembling. 
All together, they enable the generative MLLM to function effectively as a discriminative ranker within our system.

\subsection{Model Backbone}
We adopt LLaVA \cite{Liu_2024_CVPR} as the MLLM architecture, leveraging its strong performance and flexibility. It consists of three main components:
\\\texttt{LLM (Large Language Model)}: We use Mistral-7B\cite{jiang2023mistral7b}, chosen for its compatibility with industry-serving environments.
\\\texttt{Vision Encoder}: We employ ViT-Large, which provides robust visual feature extraction.
\\\texttt{Projector}: A two-layer MLP is used to align vision and language representations.
\\The training process begins with Mistral-7B, pretrained by the LLaVA team, as the initialization.
\\During fine-tuning, we follow standard next-token prediction for captioning and VQA datasets. Given a sentence that is segmented as a sequence of tokens $x = (x_1, x_2, ..., x_n)$, where $x_i$ belongs to $V$, which is the vocabulary dictionary. The joint probability of the sequence x is modeled as: 
$$p(x_1, x_2, ..., x_n) = \prod_{i=1}^{n}P(x_i|x_1,x_2,....,x_{i-1})$$
While for the finetuning on classification dataset, the task reduces to single-token prediction, where only one token represents the final classification label. The extraction of the predicted token probability is elaborated in Section~\ref{probability}.

\begin{table*}[t]
    \centering
    \small
    \begin{tabular}{lcccc}
        \toprule
        \textbf{Models} & \textbf{Prompt} & \textbf{PR-AUC} & \textbf{ROC-AUC} & \textbf{Max-F1}\\
        \midrule
        Multi-Modal Classification (X-VLM) & - & 30.79 & 65.31 & 36.81 \\
        \midrule
        LLaVA & - & 23.17 & 58.59 & 31.32 \\
        LLaVA w/ Caption & - & 28.85  &  65.88 & 36.71 \\
        \midrule
        \multirow{4}{*}{Mixed Sequential Phased Learning} & P1 & 66.96 & 87.01 & 60.64 \\
         & P2 & \underline{68.10} & \underline{87.47} & \underline{60.98} \\
         & P3 & 62.43 & 84.90 & 57.06 \\
         & P4 & 66.97 & 87.05 & 60.51 \\
        \midrule
        \multirow{4}{*}{Multi-task Learning} & P1 & 66.33 & 86.90 & 59.94 \\
         & P2 & \textbf{68.73} & \textbf{87.68} & \textbf{61.29} \\
         & P3 & 65.11 & 86.05 & 58.54 \\
         & P4 & 67.60 & 87.32 & 60.84 \\
        \bottomrule
    \end{tabular}
    \caption{Performance results (\%) across models with different training strategies and prompt designs. The top section presents results from traditional multimodal classification models. The middle section includes two zero-shot models: the first is the original LLaVA model, while the second is further fine-tuned on a captioning task. The bottom section reports results from different models fine-tuned on the classification dataset.}
    \label{tab:performance}
\end{table*}

\subsection{Training Dataset}
The training dataset consists of three parts:
\\\textbf{VQA Dataset.} A randomly subsampled dataset from LLaVA-Mix665k \cite{Liu_2024_CVPR} that is used for fine-tuning. It includes COCO, GQA, OCR-VQA, TextVQA, and VG, providing a strong foundation for visual comprehension and question-answering capabilities.
\\\textbf{Video Caption Dataset.} A high-quality caption dataset designed to provide rich contextual summaries of videos. Captions cover key aspects such as subjects, attributes, actions, and scenes. For inappropriate videos, the captions highlight potential violations based on these aspects.
\\\textbf{Classification Dataset.} This dataset is customized for moderation tasks, with each video labeled with a fine-grained issue tag and an overall label indicating whether or not action should be taken. We selected representative moderation issues and sampled the dataset according to the online traffic distribution. The dataset exactly aligns with the online data distribution after the Router.
\\In total, the dataset contains ~300k samples, with a 1:1:1 ratio across the three subsets.

\subsection{Training Strategy}
We explored two different Supervised Fine-Tuning (SFT) strategies as mentioned in the paper \cite{dong-etal-2024-abilities}. Let $D_1$, $D_2$, and $D_3$ represent the three datasets used in training. 
\\\textit{Multi-task learning.} Directly mix different fine-tuning data sources $D = \cup_{1\leq i\leq 3} D_i$ and then train on the mixed dataset. For multi-task learning, the overall training procedure is about 20 hours using 8$\times$A100 GPUs. 
\\\textit{Mixed Sequential phased learning.} The first stage is Visual Instruction Tuning. We first mix $D_1$ and $D_2$ and train to get the best epoch. Then, in the second stage, called Moderation-Oriented Supervised Fine-Tuning. We fine-tune on $D_3$ specifically for Moderation. For sequential phased training, the first phase of sequential training is about 10 hours, and the continuous training is about 10.5 hours using 8$\times$A100 GPUs.

\subsection{Transform Model Output}
\label{probability}
To make the model's output fit for actual online deployment service and more flexible to adjustments, we applied a transformation to the single token output to the actual probability. This adjustment also facilitates easier evaluation and comparison against classification models. By setting specific thresholds, we can also tune the model's behavior. Below, we present the Algorithm~\ref{alg1} illustrating this.


\begin{algorithm} 
    \small
	\renewcommand{\algorithmicrequire}{\textbf{Input:}}
	\renewcommand{\algorithmicensure}{\textbf{Output:}}
	\caption{Modified Output Pseudocode} 
	\label{alg1} 
	\begin{algorithmic}[1]
		\REQUIRE Prompt $P$, Model $M$, Tokenizer $T$
            \ENSURE Output Score $S = [p_Y, p_N]$
            \STATE \textbf{Step 1: Model Inference}
            \STATE \quad$input\_ids \gets T.tokenize(P)$
            \STATE \quad$output\_ids \gets M.generate(input\_ids)$
            \STATE \quad$logits \gets output\_ids.scores$
            \STATE \textbf{Step 2: Compute Probabilities for Answers}
            \STATE \quad$\ell_Y \gets logits[Y]$
            \STATE \quad$\ell_N \gets logits[N]$
            \STATE Compute softmax probabilities:
            \STATE \quad$p_Y \gets \frac{e^{\ell_Y}}{e^{\ell_Y} + e^{\ell_N}}$
            \STATE \quad$p_N \gets \frac{e^{\ell_N}}{e^{\ell_Y} + e^{\ell_N}}$
            \STATE \textbf{Step 3: Generate Output Score}
            \STATE \quad$S \gets [p_Y, p_N]$ \COMMENT{Final probability list}
            
            \RETURN $S$
	\end{algorithmic} 
\end{algorithm}
\begin{figure*}[b]
\centering
    \subfigure[Prompt Template 1]{
        \includegraphics[width=0.23\linewidth]{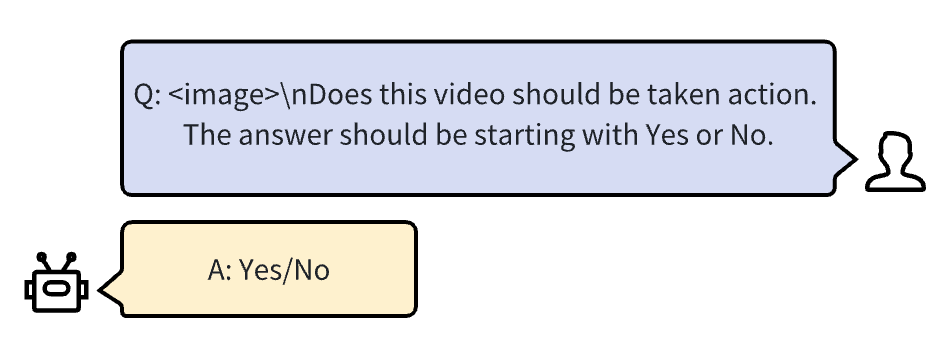}
    }
    \subfigure[Prompt Template 2]{
        \includegraphics[width=0.23\linewidth]{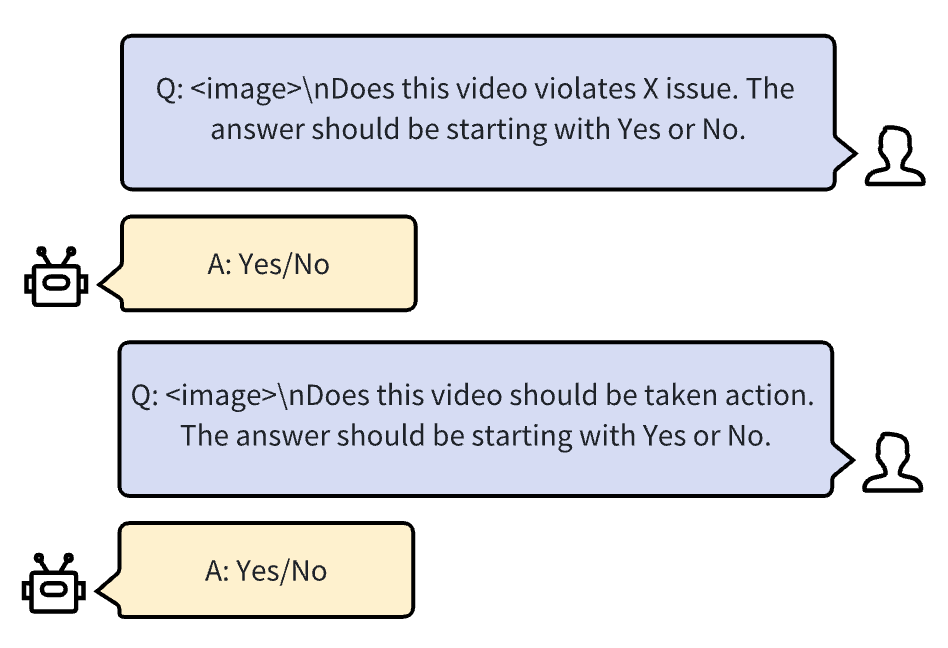}
    }
    \subfigure[Prompt Template 3]{
        \includegraphics[width=0.23\linewidth]{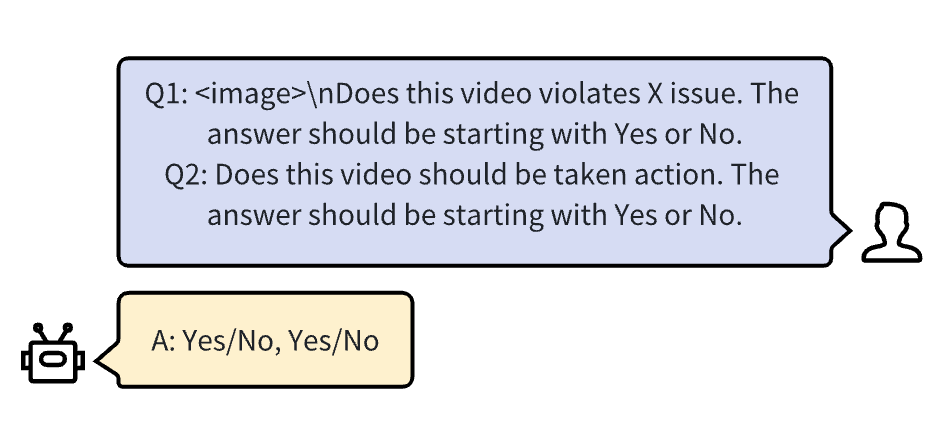}
    }
    \subfigure[Prompt Template 4]{
        \includegraphics[width=0.23\linewidth]{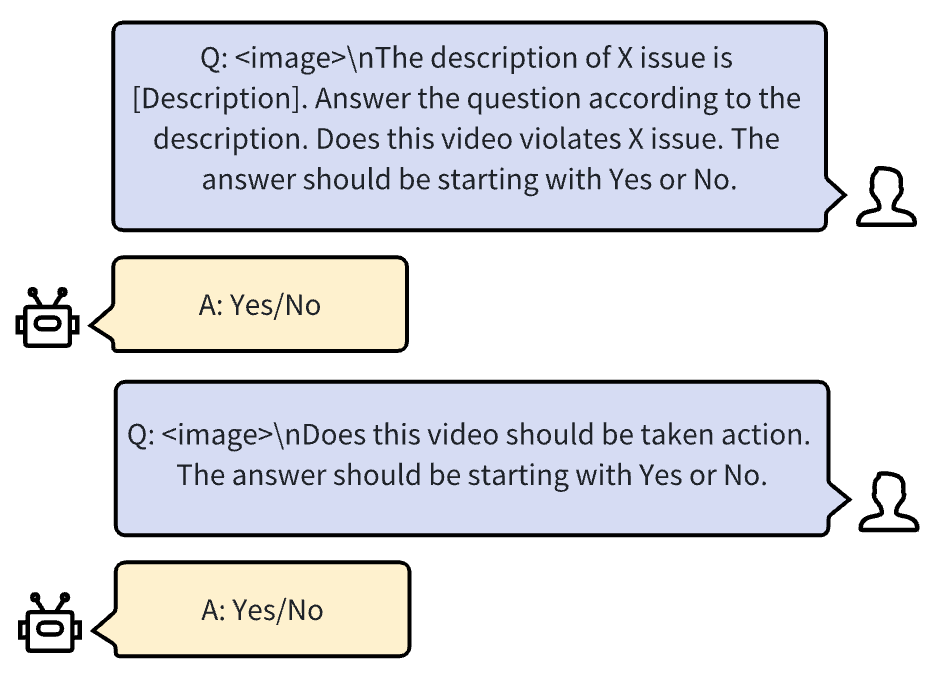}
    }
  \caption{Illustration of the four prompt templates: (a) Directly ask about the overall label, (b) Ask the fine-grained label and overall label separately, (c) Ask the fine-grained and overall labels sequentially to emphasize their relationship, and (d) Provide a definition of the fine-grained issue before asking both questions separately.}
  \label{fig:prompt}
\end{figure*}

\section{Experiments}
In this section, we introduce our experimental setup, including prompt design and adjustments to MLLM output for ranking probability. We then briefly describe the baseline models used for comparison. Finally, we present our experiments and provide a detailed analysis of the results.

\subsection{Prompt Design}
Prompt engineering plays a crucial role in optimizing MLLM performance. For our content moderation application, we designed two straightforward prompt questions, each targeting a different level of labels in the dataset. These prompts can be used independently or combined in various ways. In total, we designed four different prompt templates. (see Figure~\ref{fig:prompt} for details).

To simulate classification, we restrict the model’s output to a single-token response (Yes/No) by controlling the answer format in the training dataset. This ensures that the MLLM operates in a structured classification framework while retaining the adaptability of prompt-based reasoning.

\subsection{Baseline models}
We compare our models with two types of models:
\\\textit{Traditional Multimodal Classification Model \cite{zeng2022multigrainedvisionlanguagepretraining}.} This kind of model is widely used in modern content moderation systems. Comparison against it highlights whether our MLLM-based approach provides a performance advantage over conventional methods.
\\\textit{Zero-Shot MLLMs.} This comparison evaluates the impact of our supervised fine-tuning pipeline, demonstrating whether fine-tuned MLLMs outperform their zero-shot counterparts.

\begin{figure*}[t]
\centering
    \subfigure[LLaVA]{
        \includegraphics[width=0.31\linewidth]{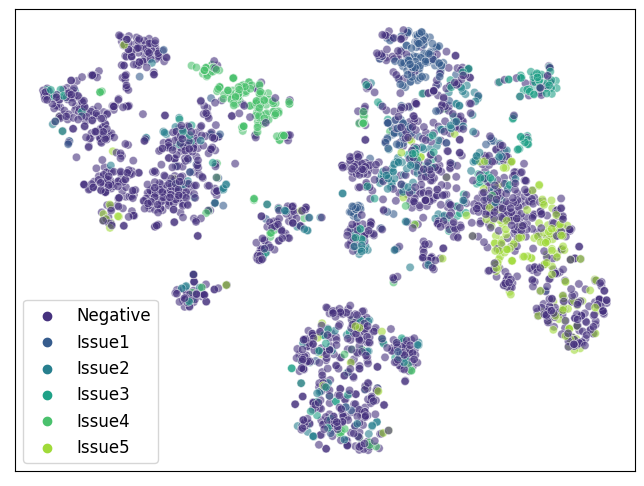}
    }
    \subfigure[LLaVA w/ Caption]{
        \includegraphics[width=0.31\linewidth]{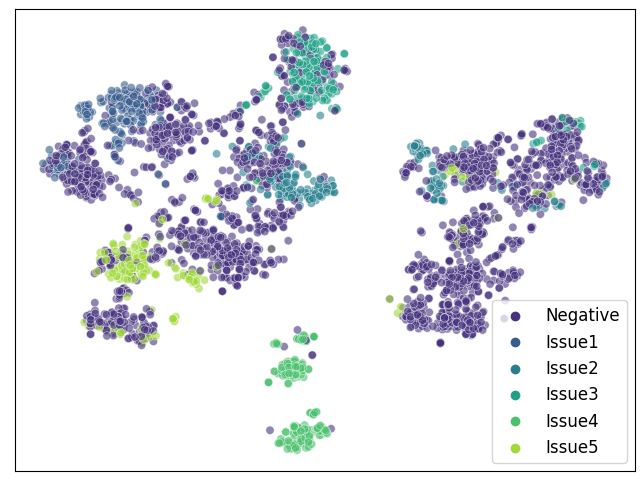}
    }
    \subfigure[Best fine-tuned model]{
        \includegraphics[width=0.31\linewidth]{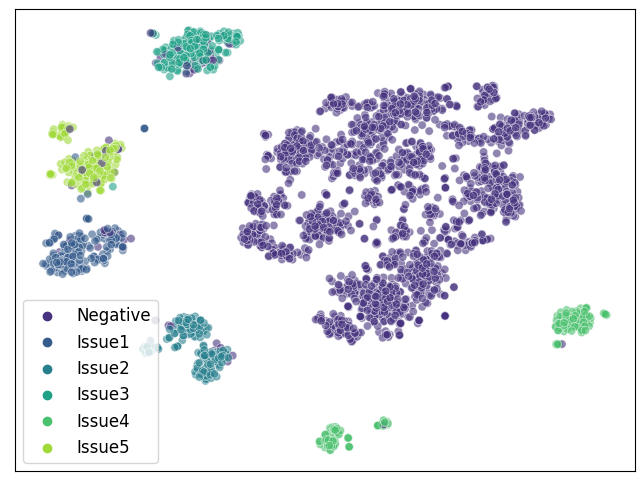}
    }
  \caption{Visualization of the embeddings extracted from the last hidden layer of each model.}
  \label{fig:visual}
\end{figure*}

\subsection{Evaluation Data and Metrics}
To ensure alignment with online data distribution, we randomly sample cases from the Router’s output and use high-quality annotators as ground truth. The final evaluation dataset consists of 50K samples. For a comprehensive performance assessment, we report PR-AUC, ROC-AUC, and Max-F1 scores. 
\subsection{Offline Evaluation Results}
From Table~\ref{tab:performance}, we may conclude the following aspects.
\\\textbf{Model Architect.} MLLM significantly outperforms traditional multimodal classification models on F1 score by 66.50\%, demonstrating its superior ability in content moderation.
\\\textbf{Supervised Fine-tuning.} Fine-tuned MLLMs outperform zero-shot models by 45.55\% in PR-AUC, confirming the effectiveness of our supervised fine-tuning pipeline.
\\\textbf{Training Strategy.} Multi-task training models consistently outperform the alternative approach across all prompts, demonstrating greater robustness. In contrast, the sequential phased training strategy is more time-efficient and flexible. It allows us to achieve nearly the same performance in significantly less time, as fine-tuning is only required in the second stage with the content moderation dataset.

\subsubsection{Ablation Study}
\textbf{Prompt Design.} Prompt design matters: Separately asking two questions yields the best performance. Single-question prompts like P1 and P3 do not provide as much information as multiple questions do. As for P2 being better than P4, it is likely because combining both labels in a single prompt introduces additional noise, confusing the final prediction of the model.
\\\textbf{Label Assemble.} We compared several widely used assembling methods to aggregate fine-grained label predictions and overall label predictions: \emph{Union Probability}, \emph{Maximum Probability}, \emph{Weighted Sum Probability}, and \emph{Bayesian Fusion Probability}\cite{chen2022multimodal}. As shown in Table~\ref{tab:label assemble}, the \emph{Weighted Sum} method achieves the highest PR-AUC, while the \emph{Union Probability} method performs best in ROC-AUC.
\begin{table}[h]
    \centering
    \small
    \begin{tabular}{lccc}
    \toprule
        \textbf{Method} & \textbf{PR-AUC} & \textbf{ROC-AUC} & \textbf{Max-F1}\\
        \midrule
        Original & 68.73 & 87.68 & \textbf{61.29} \\
        Union & 68.78 & \textbf{87.83} & 61.28 \\
        Maximum & \underline{68.79} & 87.78 & \textbf{61.29} \\
        Weighted Sum & \textbf{68.83} & 87.78 & 61.28 \\
        Bayesian Fusion & 68.67 & \underline{87.79} & 61.22 \\
    \bottomrule
    \end{tabular}
    \caption{Result(\%) of different label assemble methods.}
    \label{tab:label assemble}
\end{table}
\\\textbf{Temperature Tuning.} We experimented with different temperature values ranging from 0.2 to 0.8 to thoroughly investigate the impact of randomness on the final outcome. However, the results show that temperature does not have a big impact on model performance.
\\

\subsubsection{Visualization}
To more intuitively demonstrate the model output distribution, we extracted the final hidden layer of three models and visualized the embeddings. It is obvious that our best model draws a better decision boundary, as shown in Figure~\ref{fig:visual}.

\subsection{Online Experiment}
We deploy our cascade system online and conduct A/B experiments on 12 representative issues. We evaluate the final result on the following metrics.

\subsubsection{Action Volume and Precision Increment}
The online experiment shows an average increase of 41.27\% in action volume. Furthermore, with the addition of ranker, system-wise precision saw an improvement of 19.16\%. For a detailed breakdown of each issue, see Appendix~\ref{sec:appendix}.

\subsubsection{Resources Saving}
We observed that the router has eliminated traffic flow by 97.5\% without increasing latency in serving, which means filtering numerous compliance videos and saving resources for the ranker to better distinguish potential high-risk videos. Furthermore, compared to the traditional multimodality classification model, our MLLM uses only 2\% of the human-annotated data, significantly saving human resources.

\section{Conclusion}

In this paper, we introduced an MLLM-based cascade system for industrial-scale content moderation. Our approach demonstrated strong performance in both offline and real-world online experiments. Furthermore, our system design enables the efficient deployment of MLLMs at production scale while maintaining affordable computational costs. This solution has been successfully integrated into production systems, driving actual downstream business applications and setting a new benchmark for scalable AI-driven content moderation.

\section*{Limitations}
The current model still relies on a small amount of human-annotated data, which may introduce additional noise. Moreover, due to the limitations of the router component, the system still carries a risk of missed detection.



\bibliography{custom}

\begin{thebibliography}{32}
\providecommand{\natexlab}[1]{#1}

\bibitem[{Binh et~al.(2022)Binh, Tandon, Oinar, Liu, Durairaj, Guo, Zahabizadeh, Ilango, Tang, Morstatter et~al.}]{binh2022samba}
Le~Binh, Rajat Tandon, Chingis Oinar, Jeffrey Liu, Uma Durairaj, Jiani Guo, Spencer Zahabizadeh, Sanjana Ilango, Jeremy Tang, Fred Morstatter, et~al. 2022.
\newblock Samba: Identifying inappropriate videos for young children on youtube.
\newblock In \emph{Proceedings of the 31st ACM International Conference on Information \& Knowledge Management}, pages 88--97.

\bibitem[{Chen et~al.(2022)Chen, Shi, Ye, Mertz, Ramanan, and Kong}]{chen2022multimodal}
Yi-Ting Chen, Jinghao Shi, Zelin Ye, Christoph Mertz, Deva Ramanan, and Shu Kong. 2022.
\newblock Multimodal object detection via probabilistic ensembling.
\newblock In \emph{European Conference on Computer Vision}, pages 139--158. Springer.

\bibitem[{Chen et~al.(2024)Chen, Xu, Qi, and Guo}]{chen2024mllm}
Zhanpeng Chen, Chengjin Xu, Yiyan Qi, and Jian Guo. 2024.
\newblock {MLLM} is a strong reranker: Advancing multimodal retrieval-augmented generation via knowledge-enhanced reranking and noise-injected training.
\newblock \emph{arXiv preprint arXiv:2407.21439}.

\bibitem[{Chen et~al.(2025)Chen, Chen, Imani, and Imani}]{chen2025can}
Zhiling Chen, Hanning Chen, Mohsen Imani, and Farhad Imani. 2025.
\newblock Can multimodal large language models be guided to improve industrial anomaly detection?
\newblock \emph{arXiv preprint arXiv:2501.15795}.

\bibitem[{Christiano et~al.(2017)Christiano, Leike, Brown, Martic, Legg, and Amodei}]{deepreinforcement}
Paul~F. Christiano, Jan Leike, Tom~B. Brown, Miljan Martic, Shane Legg, and Dario Amodei. 2017.
\newblock Deep reinforcement learning from human preferences.
\newblock In \emph{Proceedings of the 31st International Conference on Neural Information Processing Systems}, NIPS'17, page 4302–4310, Red Hook, NY, USA. Curran Associates Inc.

\bibitem[{Das et~al.(2023)Das, Raj, Saha, Mathew, Gupta, and Mukherjee}]{das2023hatemm}
Mithun Das, Rohit Raj, Punyajoy Saha, Binny Mathew, Manish Gupta, and Animesh Mukherjee. 2023.
\newblock Hatemm: A multi-modal dataset for hate video classification.
\newblock In \emph{Proceedings of the International AAAI Conference on Web and Social Media}, volume~17, pages 1014--1023.

\bibitem[{DeepSeek-AI(2025{\natexlab{a}})}]{deepseekai2025deepseekr1incentivizingreasoningcapability}
DeepSeek-AI. 2025{\natexlab{a}}.
\newblock \href {https://arxiv.org/abs/2501.12948} {Deepseek-r1: Incentivizing reasoning capability in llms via reinforcement learning}.
\newblock \emph{Preprint}, arXiv:2501.12948.

\bibitem[{DeepSeek-AI(2025{\natexlab{b}})}]{deepseekai2025deepseekv3technicalreport}
DeepSeek-AI. 2025{\natexlab{b}}.
\newblock \href {https://arxiv.org/abs/2412.19437} {Deepseek-v3 technical report}.
\newblock \emph{Preprint}, arXiv:2412.19437.

\bibitem[{Dong et~al.(2024)Dong, Yuan, Lu, Li, Xue, Liu, Wang, Yuan, Zhou, and Zhou}]{dong-etal-2024-abilities}
Guanting Dong, Hongyi Yuan, Keming Lu, Chengpeng Li, Mingfeng Xue, Dayiheng Liu, Wei Wang, Zheng Yuan, Chang Zhou, and Jingren Zhou. 2024.
\newblock \href {https://doi.org/10.18653/v1/2024.acl-long.12} {How abilities in large language models are affected by supervised fine-tuning data composition}.
\newblock In \emph{Proceedings of the 62nd Annual Meeting of the Association for Computational Linguistics (Volume 1: Long Papers)}, pages 177--198, Bangkok, Thailand. Association for Computational Linguistics.

\bibitem[{Jiang et~al.(2023)Jiang, Sablayrolles, Mensch, Bamford, Chaplot, de~las Casas, Bressand, Lengyel, Lample, Saulnier, Lavaud, Lachaux, Stock, Scao, Lavril, Wang, Lacroix, and Sayed}]{jiang2023mistral7b}
Albert~Q. Jiang, Alexandre Sablayrolles, Arthur Mensch, Chris Bamford, Devendra~Singh Chaplot, Diego de~las Casas, Florian Bressand, Gianna Lengyel, Guillaume Lample, Lucile Saulnier, Lélio~Renard Lavaud, Marie-Anne Lachaux, Pierre Stock, Teven~Le Scao, Thibaut Lavril, Thomas Wang, Timothée Lacroix, and William~El Sayed. 2023.
\newblock \href {https://arxiv.org/abs/2310.06825} {Mistral 7b}.
\newblock \emph{Preprint}, arXiv:2310.06825.

\bibitem[{Liang et~al.(2025)Liang, Shi et~al.}]{liang2025embedding}
Hanzhong Liang, Jinghao Shi, et~al. 2025.
\newblock Embedding-based retrieval in multi-modal content moderation.
\newblock In \emph{Proceedings of the 48th International ACM SIGIR Conference on Research and Development in Information Retrieval (SIGIR)}, Padua, Italy.
\newblock To appear.

\bibitem[{Liu et~al.(2024{\natexlab{a}})Liu, Sun, Sun, Dong, and Xiong}]{liu2024agentps}
Gorden Liu, Yu~Sun, Ruixiao Sun, Xin Dong, and Hongyu Xiong. 2024{\natexlab{a}}.
\newblock Agentps: Agentic process supervision for multi-modal content quality assurance through multi-round qa.
\newblock \emph{arXiv preprint arXiv:2412.15251}.

\bibitem[{Liu et~al.(2024{\natexlab{b}})Liu, Li, Li, and Lee}]{Liu_2024_CVPR}
Haotian Liu, Chunyuan Li, Yuheng Li, and Yong~Jae Lee. 2024{\natexlab{b}}.
\newblock Improved baselines with visual instruction tuning.
\newblock In \emph{Proceedings of the IEEE/CVF Conference on Computer Vision and Pattern Recognition (CVPR)}, pages 26296--26306.

\bibitem[{Liu et~al.(2023)Liu, Li, Wu, and Lee}]{NEURIPS2023_6dcf277e}
Haotian Liu, Chunyuan Li, Qingyang Wu, and Yong~Jae Lee. 2023.
\newblock \href {https://proceedings.neurips.cc/paper_files/paper/2023/file/6dcf277ea32ce3288914faf369fe6de0-Paper-Conference.pdf} {Visual instruction tuning}.
\newblock In \emph{Advances in Neural Information Processing Systems}, volume~36, pages 34892--34916. Curran Associates, Inc.

\bibitem[{Ma et~al.(2023)Ma, Zhang, Fu, Zhao, and Wu}]{ma2023adapting}
Huan Ma, Changqing Zhang, Huazhu Fu, Peilin Zhao, and Bingzhe Wu. 2023.
\newblock Adapting large language models for content moderation: Pitfalls in data engineering and supervised fine-tuning.
\newblock \emph{arXiv preprint arXiv:2310.03400}.

\bibitem[{Mitra et~al.(2025)Mitra, Huang, Chai, Lin, Arbelle, Feris, Karlinsky, Darrell, Ramanan, and Herzig}]{mitra2025sparseattentionvectorsgenerative}
Chancharik Mitra, Brandon Huang, Tianning Chai, Zhiqiu Lin, Assaf Arbelle, Rogerio Feris, Leonid Karlinsky, Trevor Darrell, Deva Ramanan, and Roei Herzig. 2025.
\newblock \href {https://arxiv.org/abs/2412.00142} {Sparse attention vectors: Generative multimodal model features are discriminative vision-language classifiers}.
\newblock \emph{Preprint}, arXiv:2412.00142.

\bibitem[{Mullick et~al.(2023)Mullick, Bhambhani, Sinha, Mathur, Gupta, and Shah}]{mullick2023content}
Sankha~Subhra Mullick, Mohan Bhambhani, Suhit Sinha, Akshat Mathur, Somya Gupta, and Jidnya Shah. 2023.
\newblock Content moderation for evolving policies using binary question answering.
\newblock In \emph{Proceedings of the 61st Annual Meeting of the Association for Computational Linguistics (Volume 5: Industry Track)}, pages 561--573.

\bibitem[{notAI.tech(2024)}]{nudenet}
notAI.tech. 2024.
\newblock \href {https://github.com/notAI-tech/NudeNet} {Nudenet: lightweight nudity detection}.

\bibitem[{OpenAI(2024)}]{openai2024gpt4technicalreport}
OpenAI. 2024.
\newblock \href {https://arxiv.org/abs/2303.08774} {{GPT-4 Technical Report}}.
\newblock \emph{Preprint}, arXiv:2303.08774.

\bibitem[{Ouyang et~al.(2022)Ouyang, Wu, Jiang, Almeida, Wainwright, Mishkin, Zhang, Agarwal, Slama, Ray, Schulman, Hilton, Kelton, Miller, Simens, Askell, Welinder, Christiano, Leike, and Lowe}]{NEURIPS2022_b1efde53}
Long Ouyang, Jeffrey Wu, Xu~Jiang, Diogo Almeida, Carroll Wainwright, Pamela Mishkin, Chong Zhang, Sandhini Agarwal, Katarina Slama, Alex Ray, John Schulman, Jacob Hilton, Fraser Kelton, Luke Miller, Maddie Simens, Amanda Askell, Peter Welinder, Paul~F Christiano, Jan Leike, and Ryan Lowe. 2022.
\newblock \href {https://proceedings.neurips.cc/paper_files/paper/2022/file/b1efde53be364a73914f58805a001731-Paper-Conference.pdf} {Training language models to follow instructions with human feedback}.
\newblock In \emph{Advances in Neural Information Processing Systems}, volume~35, pages 27730--27744. Curran Associates, Inc.

\bibitem[{Pang et~al.(2024)Pang, Zhou, Zhou, and Wang}]{pang-etal-2024-phased}
Wei Pang, Chuan Zhou, Xiao-Hua Zhou, and Xiaojie Wang. 2024.
\newblock \href {https://doi.org/10.18653/v1/2024.findings-acl.341} {Phased instruction fine-tuning for large language models}.
\newblock In \emph{Findings of the Association for Computational Linguistics: ACL 2024}, pages 5735--5748, Bangkok, Thailand. Association for Computational Linguistics.

\bibitem[{Pareja et~al.(2024)Pareja, Nayak, Wang, Killamsetty, Sudalairaj, Zhao, Han, Bhandwaldar, Xu, Xu, Han, Inglis, and Srivastava}]{pareja2024unveilingsecretrecipeguide}
Aldo Pareja, Nikhil~Shivakumar Nayak, Hao Wang, Krishnateja Killamsetty, Shivchander Sudalairaj, Wenlong Zhao, Seungwook Han, Abhishek Bhandwaldar, Guangxuan Xu, Kai Xu, Ligong Han, Luke Inglis, and Akash Srivastava. 2024.
\newblock \href {https://arxiv.org/abs/2412.13337} {Unveiling the secret recipe: A guide for supervised fine-tuning small llms}.
\newblock \emph{Preprint}, arXiv:2412.13337.

\bibitem[{Shi et~al.(2024)Shi, Shen, Zhao, Wang, Wen, Wang, Wu, and Zhang}]{shi2024cpfd}
Jinghao Shi, Xiang Shen, Kaili Zhao, Xuedong Wang, Vera Wen, Zixuan Wang, Yifan Wu, and Zhixin Zhang. 2024.
\newblock {CPFD:} confidence-aware privileged feature distillation for short video classification.
\newblock In \emph{Proceedings of the 33rd ACM International Conference on Information and Knowledge Management}, pages 4866--4873.

\bibitem[{Stiennon et~al.(2020)Stiennon, Ouyang, Wu, Ziegler, Lowe, Voss, Radford, Amodei, and Christiano}]{learningtosummaize}
Nisan Stiennon, Long Ouyang, Jeff Wu, Daniel~M. Ziegler, Ryan Lowe, Chelsea Voss, Alec Radford, Dario Amodei, and Paul Christiano. 2020.
\newblock Learning to summarize from human feedback.
\newblock In \emph{Proceedings of the 34th International Conference on Neural Information Processing Systems}, NIPS '20, Red Hook, NY, USA. Curran Associates Inc.

\bibitem[{Wu et~al.(2024)Wu, Zhao, Cao, Xu, Jiang, Wang, Li, Hu, Qin, and Fu}]{wu2024icm}
Mengyang Wu, Yuzhi Zhao, Jialun Cao, Mingjie Xu, Zhongming Jiang, Xuehui Wang, Qinbin Li, Guangneng Hu, Shengchao Qin, and Chi-Wing Fu. 2024.
\newblock {ICM-Assistant:} instruction-tuning multimodal large language models for rule-based explainable image content moderation.
\newblock \emph{arXiv preprint arXiv:2412.18216}.

\bibitem[{Xu et~al.(2025)Xu, Shu, Mei, Xie, Fernando, and Tang}]{xu2025fedmllmfederatedfinetuningmllm}
Binqian Xu, Xiangbo Shu, Haiyang Mei, Guosen Xie, Basura Fernando, and Jinhui Tang. 2025.
\newblock \href {https://arxiv.org/abs/2411.14717} {Fedmllm: Federated fine-tuning mllm on multimodal heterogeneity data}.
\newblock \emph{Preprint}, arXiv:2411.14717.

\bibitem[{Ye et~al.(2023)Ye, Sikka, Atwell, Hassan, Divakaran, and Alikhani}]{ye-etal-2023-multilingual}
Meng Ye, Karan Sikka, Katherine Atwell, Sabit Hassan, Ajay Divakaran, and Malihe Alikhani. 2023.
\newblock \href {https://doi.org/10.18653/v1/2023.eacl-main.276} {Multilingual content moderation: A case study on {R}eddit}.
\newblock In \emph{Proceedings of the 17th Conference of the European Chapter of the Association for Computational Linguistics}, pages 3828--3844, Dubrovnik, Croatia. Association for Computational Linguistics.

\bibitem[{Yu et~al.(2025)Yu, Li, Wu, Wen, Deng, and Xiong}]{yu2025usmunbiasedsurveymodeling}
Chenghui Yu, Peiyi Li, Haoze Wu, Yiri Wen, Bingfeng Deng, and Hongyu Xiong. 2025.
\newblock \href {https://arxiv.org/abs/2412.10674} {Usm: Unbiased survey modeling for limiting negative user experiences in recommendation systems}.
\newblock \emph{Preprint}, arXiv:2412.10674.

\bibitem[{Yuan et~al.(2024)Yuan, Yu, Mittal, Hall, Sajeev, and Chen}]{yuan2024rethinking}
Jialin Yuan, Ye~Yu, Gaurav Mittal, Matthew Hall, Sandra Sajeev, and Mei Chen. 2024.
\newblock Rethinking multimodal content moderation from an asymmetric angle with mixed-modality.
\newblock In \emph{Proceedings of the IEEE/CVF Winter Conference on Applications of Computer Vision}, pages 8532--8542.

\bibitem[{Zeng et~al.(2022)Zeng, Zhang, and Li}]{zeng2022multigrainedvisionlanguagepretraining}
Yan Zeng, Xinsong Zhang, and Hang Li. 2022.
\newblock \href {https://arxiv.org/abs/2111.08276} {Multi-grained vision language pre-training: Aligning texts with visual concepts}.
\newblock \emph{Preprint}, arXiv:2111.08276.

\bibitem[{Zhang et~al.(2024)Zhang, Unell, Wang, Ghosh, Su, Schmidt, and Yeung-Levy}]{zhang2024visually}
Yuhui Zhang, Alyssa Unell, Xiaohan Wang, Dhruba Ghosh, Yuchang Su, Ludwig Schmidt, and Serena Yeung-Levy. 2024.
\newblock Why are visually-grounded language models bad at image classification?
\newblock \emph{arXiv preprint arXiv:2405.18415}.

\bibitem[{Zhou et~al.(2024)Zhou, He, Ke, Zhu, Gutierrez~Basulto, and Pan}]{zhou-etal-2024-empirical}
Xiongtao Zhou, Jie He, Yuhua Ke, Guangyao Zhu, Victor Gutierrez~Basulto, and Jeff Pan. 2024.
\newblock \href {https://doi.org/10.18653/v1/2024.findings-acl.598} {An empirical study on parameter-efficient fine-tuning for {M}ulti{M}odal large language models}.
\newblock In \emph{Findings of the Association for Computational Linguistics: ACL 2024}, pages 10057--10084, Bangkok, Thailand. Association for Computational Linguistics.

\end{thebibliography}

\appendix

\section{Detailed Experiment Result}
\label{sec:appendix}
This is a detailed breakdown of the volume increase for each issue.
\begin{table}[h]
    \centering
    \small
    \begin{tabular}{cc}
        \toprule
        \textbf{Issue} & \textbf{Action Volume Increase (\%)} \\
        \midrule
        \textbf{1}  & 47.07 \\
        \textbf{2}  & 59.96 \\
        \textbf{3}  & 45.18 \\
        \textbf{4}  & 27.64 \\
        \textbf{5}  & 22.03 \\
        \textbf{6}  & 36.04 \\
        \textbf{7}  & 41.78 \\
        \textbf{8}  & 65.62 \\
        \textbf{9}  & 26.11 \\
        \textbf{10} & 63.31 \\
        \textbf{11} & 29.66 \\
        \textbf{12} & 30.88 \\
        \midrule
        \textbf{Average} & 41.27 \\
        \bottomrule
    \end{tabular}
    \caption{Action Volume Increase for Each Issue}
    \label{tab:online_result}
\end{table}

\end{document}